\documentclass{article}

     \usepackage[preprint]{neurips_2020}

\usepackage[utf8]{inputenc} 
\usepackage[T1]{fontenc}    
\usepackage{url}            
\usepackage{booktabs}       
\usepackage{amsfonts}       
\usepackage{nicefrac}       
\usepackage{microtype}      

\usepackage{graphicx}
\usepackage{amsmath}
\usepackage{amsthm}
\usepackage{booktabs}
\usepackage{algorithm}
\usepackage{algorithmic}
\usepackage{amsfonts}
\usepackage{subfigure}

\title{IC-Network: Efficient Structure for Convolutional Neural Networks}

\author{
	Junyi An$^1$
	Fengshan Liu$^1$
	Jian zhao$^2$
	Furao shen$^1$\footnote{Contact Author}\\
	$^1$Department of Computer Science and Technology, Nanjing University, Nanjing, China\\
	$^2$School of Electronic Science and Engineering, Nanjing University, Nanjing, China\\
	\texttt{ \{junyian, liufengshan\}@smail.nju.edu.cn }\\
	\texttt{ \{frshen, jianzhao\}@nju.edu.cn }
}

\begin{document}

\maketitle

\begin{abstract}
  Neural networks have been widely used, and most networks achieve excellent performance by stacking certain types of basic units. Compared to increasing the depth and width of the network, designing more effective basic units has become an important research topic. Inspired by the elastic collision model in physics, we present a universal structure that could be integrated into the existing neural networks to speed up the training process and increase their generalization abilities. We term it the ``Inter-layer Collision" (IC) structure. We built two kinds of basic computational units (IC layer and IC block) that compose the convolutional neural networks (CNNs) by combining the IC structure with the convolution operation. Compared to traditional convolution, both of the proposed computational units have a stronger non-linear representation ability and can filter features useful for a given task. Using these computational units to build networks, we bring significant improvements in performance for existing state-of-the-art CNNs. On the ImageNet experiment, we integrate the IC block into ResNet-50 and reduce the top-1 error from $22.85\%$ to $21.49\%$, which also exceeds the top-1 error of ResNet-100 ($21.75\%$).
\end{abstract}


\section{Introduction}

	Convolutional neural networks (CNNs) have made great achievements in computer vision. The success of Alexnet \cite{krizhevsky2012imagenet} and VGGNet \cite{simonyan2014very} shows the superiority of deep networks, leading to a trend of building larger and deeper networks. However, this method is not efficient in improving network performance. On the one hand, increasing the depth and width brings a huge computational burden, and causes a series of problems, such as vanishing/exploding gradients, degradation, over-parameterization, etc. On the other hand, because the relationship between hyper-parameters is complicated, the increased number of hyper-parameters makes it more difficult to design networks. Therefore, the focus of research in recent years has gradually shifted to improving the representation capabilities of basic network units in order to design more efficient architectures.

	The convolutional layer is a basic unit of CNN. By stacking a series of convolutional layers together with non-linear activation layers, CNNs are able to produce image representations that capture abstract features and cover global theoretical receptive fields. For each convolutional layer, the filter can capture a certain type of feature in the input through a sliding window. However, not all features contribute to a given task. Recent studies have shown that by emphasizing informative features and suppressing less useful ones, the network can obtain more powerful representation capabilities \cite{bell2016inside} \cite{hu2018squeeze}. Besides, there is evidence that introducing non-linear factors in the convolutional layer can improve the generalization ability of network \cite{wang2019kervolutional} \cite{zoumpourlis2017non}. Some studies attempt to combine kernel methods with convolutional layers to improve generalization abilities, but it may make the networks too complicated and introduce a high amount of calculation. We argue that the convolutional layer can also benefit from a simple non-linear representation, and propose a method of introducing a non-linear operation in the convolutional layer. The proposed method can filter different input features and strengthen the representation ability of each convolutional layer.

	We first propose a neural network structure that mimics the transmission of information in a physical system, which we term the ``Inter-layer Collision" (IC) structure. This structure is inspired by the elastic collision model, and the flow of information can be viewed as a collision between neurons of two adjacent layers. We further analyze the generalization ability of the IC structure and prove that it can provide a stronger non-linear representation than the traditional neural network with the same number of learnable parameters. The detail will be introduced in Section 3.

	By combining the IC structure with the convolution operation, we build two basic computational units, which are the IC layer and the IC block. The IC layer is built on a one convolutional layer which makes it a universal structure, while the IC block is built on a block structure so that it is more suitable for building deep CNNs. Both computational units help to strengthen the non-linear representation abilities of CNNs while retain the advantages of convolution such as shared weights and low computational complexity.

	
	In ImageNet experiments, IC-ResNet-50 using IC block achieves the 10-crop top-1 error of $21.49\%$, exceeding ResNet-50 by $1.36\%$. Remarkably, this result also exceeds the error rate achieved by ResNet-101 ($21.75\%$ error) with a less computational burden. Besides, we also experimentally show that the IC structure is also effective on some other modern networks.

	\section{Related Work}

	To address the problems in training deep CNNs, a series of techniques have been suggested in the last few years. For instance, using the rectified linear unit (ReLU) \cite{nair2010rectified} as the non-linear operator is popular in CNNs since it mitigates the vanishing gradient by expanding the non-zero range of the gradient. In contrast, Batch Normalization (BN) \cite{ioffe2015batch} eases the exploding gradient problem by standardizing the mean and variance of the outputs to be the same, which transforms the output from a saturated region to an unsaturated one \cite{santurkar2018does}. However, the performance of deep networks may still be worse than a network with fewer layers because of degradation \cite{he2015convolutional}, a type of networks termed Residual Networkts (ResNets) \cite{he2016deep} were proposed to solve this problem by using skip connection. ResNet also proposed the basic and bottleneck blocks that were widely researched to further improve the performance. For instance, ResNext \cite{xie2017aggregated} designed a multi-branch architecture based on the bottleneck block, and attained a lower error rate than ResNet. 
	
	It is of vital importance to make the unit in CNNs more effective. On the one hand, the blocks regarded as a basic computational unit eases the difficulty of manually designing the network, and improving it can affect the performance of the entire network. On the other hand, Neural Architecture Search (NAS), a class of methods using algorithms to automatically search the optimal network \cite{liu2018darts}, \cite{zoph2018learning}, have achieved higher accuracy than artificial architecture on image classification. However, the modern NAS methods mainly used reinforcement learning, and still needed to define the basic unit in search space. Squeeze-and-Excitation Networks (SENet) \cite{hu2018squeeze} proposed SE blocks based on ResNet, which attained great performance both on artificial network and NAS method \cite{tan2019mnasnet}. The central idea of SENet was modeling the interdependencies between the features. 

	Adding non-linearity to convolutions was also a modern method to improve performance. Some research tried to integrate the kernel methods into convolution \cite{cohen2016deep}. In this way, the networks had a better generalization ability without disrupting the structure of the original network. However, kernel methods usually extracted features in high dimensional space, resulting in high computational complexity. To solve this problem, \cite{wang2019kervolutional} proposed Kervolutional to bypass the explicit calculation of the high dimensional features via the kernel trick.

	\section{Inter-layer Collision Structure}\label{sec3}
	
	The IC structure is inspired by the elastic collision model. In this section, we first describe the physical scene that the IC structure imitates. Then, we introduce the basic form of IC structure in neuron networks and show our analysis of why the IC structure can improve the generalization ability of the networks. Finally, applications on CNN will be introduced in detail.
	
	\subsection{The physical collision scene}

	Considering two objects in a 1D space where objects could only move to the left $(-)$ or to the right $(+)$, the mass of them are $m_{1}$ and $m_{2}$, respectively. Initially, $m_{1}$ lies to the left of $m_{2}$ and they both have a velocity of zero. When $m_{1}$ is given a speed $v_{1}$ toward $m_{2}$, according to the laws of energy conservation and momentum conservation, the velocity of $m_{1}$ and $m_{2}$ after collision are:
	\begin{equation}
		v_{1}' = \frac{m_{1}-m_{2}}{m_{1}+m_{2}}v_{1}, v_{2}' = \frac{2m_{1}}{m_{1}+m_{2}}v_{1}. 
	\end{equation}
	According to above analysis, the object $m_{2}$ will always move to the right ($+$) while the moving direction of the object $m_{1}$ depending on the quantitative relationship between $m_{1}$ and $m_{2}$. We can treat this model as an information transmitting system where we only care about the overall information flowing out of the system given the input information. In this case, the input information is $v_{1}$ and the output information is $v'' = v_{1}'' + v_{2}''$. Here $v_{1}''$ and $v_{2}''$ are given by:
	\begin{equation}
		\label{eq2}
		v_{1}'' = \sigma \left((w-1)v_{1} \right),  v_{2}'' = wv_{1}, 
	\end{equation}
	where $w$ denotes the coefficient $\frac{2m_{1}}{m_{1}+m_{2}}$, $\sigma$ denotes the ReLU function which is used to get the right component $(+)$ of $v_{1}'$. We notice that $w$ acts as a parameter controlling how much information could be transmitted. In the way, if we make $w$ learnable like what we do in machine learning and add this structure into the neural network framework, the system can be tuned to transmit useful information to the subsequent layers. Based on this assumption, we build the basic IC computational unit detailed in the next section.

	\subsection{The basic IC computational unit}

	By mimicking the relationship of two objects in the elastic collision model, we design the network with multiple neurons. Considering a two-layer structure with only input and output neurons, we replace the ${v_{1}, v''}$ with ${x, y}$ to denote the input and output respectively. Each output $y_{j}$ is defined by:
	\begin{equation}
		\label{eq3}
		y_{j} = f \left( \sum_{i=1}^{N} \ w_{ij}x_{i} + \sigma \left(\sum_{i=1}^{N} \ (w_{ij} - 1)x_{i} \right) \right), j \in [1, M], 
	\end{equation}
	where ${N,M}$ is the dimension of the input and output. $f$ is used to denote any activation functions. To simplify the notation, we omit the bias term in this paper. Intuitively, Eq. 3 can improve the overall generalization ability of the neural network by replacing the original linear transformation with a non-linear one. The term $H = \sum_{i=1}^{N} \ (w_{ij} - 1)x_{i}$ represents a hyperplane in a $N$-dimensional Euclidean space, which divides Eq. 3 as follows:
	\begin{equation}
		\label{eq4}
		 y_{j} =
	\begin{cases}
		f \left( 2\sum_{i=1}^{N} \ w_{ij}x_{i} - \sum_{i=1}^{N} \ x_{i} \right) & \text{if } H \ge 0 \\
		f \left(\sum_{i=1}^{N} \ w_{ij}x_{i} \right) & \text{if } H < 0
    \end{cases}.
	\end{equation}
	Specially, if $f$ is the ReLU activation function and the input dimension is $2$, the traditional ReLU neuron divides the space into two parts, while Eq. 3 can divide it into three parts, as shown in Fig. 1(a, b). Furthermore, we use the XOR problem to explain their difference. Since the decision boundary of the ReLU neuron is linear, it is clear that the XOR problem (linear inseparability) can not be solved by using a single ReLU neuron. Fig. 1(c) shows one solution of Eq. 3. Although the boundary is also linear, four points are divided into three spaces. Point $a$ and $c$ share the same representation because they lie in the same region and we can make the representation of point $b$ and $d$ as similar as possible by adjusting the weights. 

	\begin{figure}[t]
		\centering
		\subfigure[]{
		\centering
		\includegraphics[height=2.6cm, width=2.6cm]{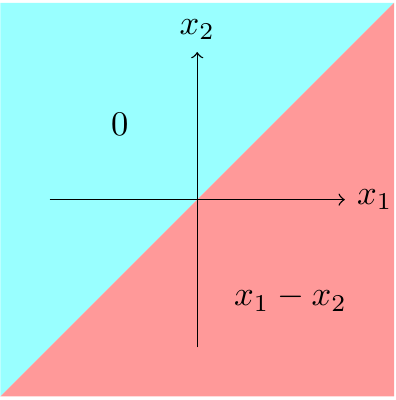}
	}
	\hspace{1cm}
	\subfigure[]{
		\centering
		\includegraphics[height=2.6cm, width=2.6cm]{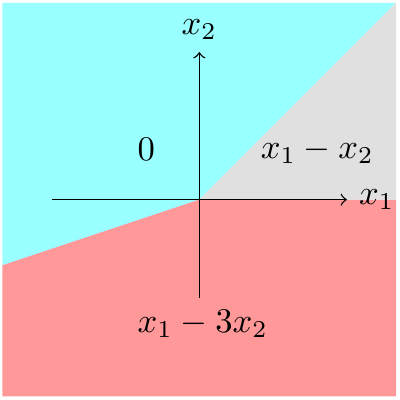}
	}
	\hspace{1cm}
	\subfigure[]{
		\centering
		\includegraphics[height=2.6cm, width=2.6cm]{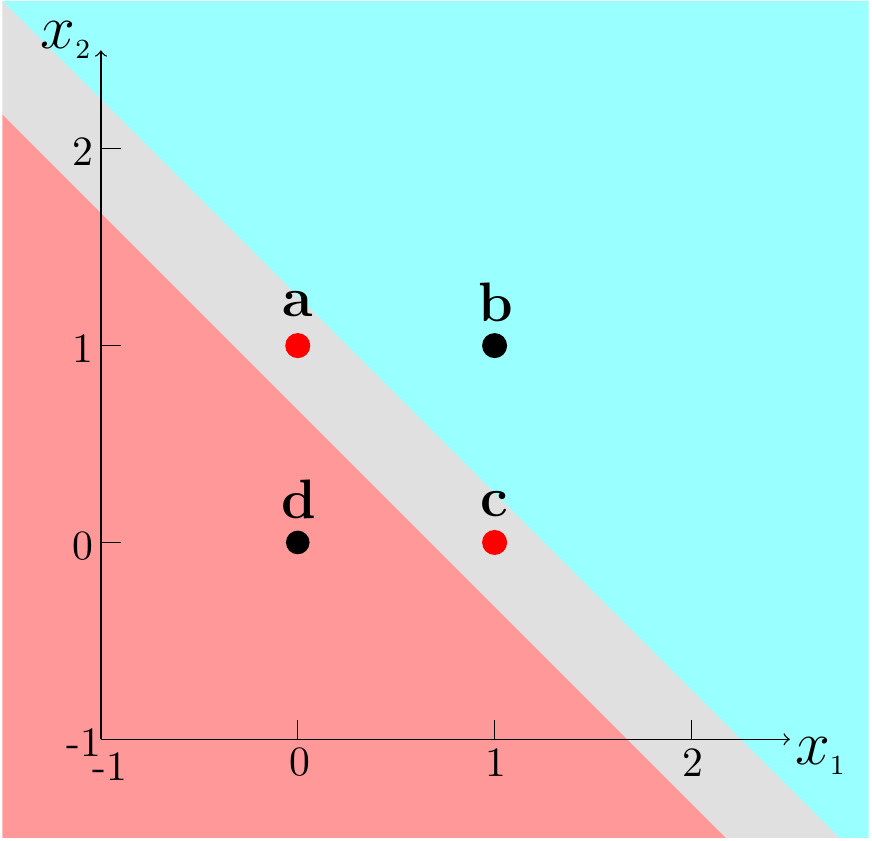}
	}
	\caption{The value of $f(x_{1},x_{2})$ given $x_{1}, x_{2}$. (a): $f(x_{1},x_{2})=\sigma (x_{1}-x_{2})$. (b): $f(x_{1},x_{2})=\sigma (x_{1}-x_{2}+\sigma(-2x_{2}))$. (c): $f(x_{1},x_{2})=\sigma (w_{1}x_{1}+w_{2}x_{2}+b_{1}+\sigma((w_{1}-1)x_{1}+(w_{2}-1)x_{2}+b_{2}))$. Here $w_{1} = w_{2} = 0.2805$. $b_{1} = -0.3506$ and $b_{2} = 0.6463$ are used to shift the boundary across the whole space. }
	\label{fig1}
	\end{figure}

	To add more representation flexibility to the hyperplane $H=0$, we further relax the constant $1$ in Eq. 3 to a learnable value $w'_{j}, j \in [0, 1, \dots, M]$. The standard IC structure is defined by:
	\begin{equation}
		\label{eq5}
		\begin{aligned}
		y_{j} = f \left(\sum_{i=1}^{N} \ w_{ij}x_{i} + \sigma \left(\sum_{i=1}^{N} \ (w_{ij} - w'_{j})x_{i} \right) \right) = f \left(\sum_{i=1}^{N} \ w_{ij}x_{i} + \sigma \left(\sum_{i=1}^{N} \ w_{ij}x_{i} - w'_{j}x_{sum} \right) \right) ,
		\end{aligned}
	\end{equation}
	where $x_{sum}$ denotes the sum of input. The learnable weight $w'_{j}$ makes it possible that $ \sigma(H) = 0 $ whatever $w_{ij}$ is, retaining the representation abilities of traditional neurons. When $ \sigma(H) > 0$, $w'_{j}$  can provide more flexible decision boundaries, improving the overall generalization abilities.
	\paragraph{Theorem 3.1} By adjusting $w'$, the hyperplane $\sum_{i=1}^{N} \ (w_{i} - w')x_{i} = 0$ can be rotated $\pi$ around the cross product of vector $\mathbf{W}$ and $\mathbf{I}$ when two vectors are linearity independent. Here $\mathbf{W} = (w_{1},w_{2},\dots,w_{N})^{T}$ and $\mathbf{I} = (\underbrace{1,1,\cdots,1}_{N})^{T}$.
	\begin{proof}
		The normal vector of $\sum_{i=1}^{N} \ (w_{i} - w')x_{i} = 0$ is given by $\mathbf{H} = (w_{1}-w',w_{2}-w',\dots,w_{N}-w')^{T}$. Consider the angle between $\mathbf{H}$ and $\mathbf{I}$:
		\begin{equation}
			\label{eq6}
			\cos(\theta) = \frac{\mathbf{H}^{T} \cdot \mathbf{I}}{|\mathbf{H}| |\mathbf{I}|}
			 = \frac{\mathbf{W}^{T} \cdot \mathbf{I} - Nw'}{\sqrt{N|\mathbf{W}|^{2} - 2Nw'\mathbf{W}^{T} \cdot \mathbf{I} + N^{2}w'^{2}}}. 
		\end{equation}
		When $\mathbf{W}^{T} \cdot \mathbf{I} \ge Nw'$: 
		\begin{equation}
			\label{eq7}
			\cos(\theta) = \sqrt{1 + \frac{(\mathbf{W}^{T} \cdot \mathbf{I})^{2} - N|\mathbf{W}|^{2}}{N|\mathbf{W}|^{2} - 2Nw'\mathbf{W}^{T} \cdot \mathbf{I} + N^{2}w'^{2}}}.
		\end{equation}
		Eq. 6 is a continuous subtractive function, and $\cos(\theta) \in [0,1)$ when $w' \in (-\infty,\frac{\mathbf{W}^{T} \cdot \mathbf{I}}{N}]$. Similarly, $\cos(\theta) \in (-1,0]$ when $w' \in [\frac{\mathbf{W}^{T} \cdot \mathbf{I}}{N},\infty)$. Therefore, we get $\theta \in (0,\pi)$ when $w' \in (-\infty,\infty)$. It is clear that direction of rotation axis is same as the cross product of $\mathbf{W}$ and $\mathbf{I}$.
	\end{proof}
	Theorem 3.1 implies that $\sum_{N}^{i=1} \ (w_{i} - w')x_{i} = 0$ can almost represent the whole hyperplanes parallel to the cross product of $\mathbf{W}$ and $\mathbf{I}$, providing flexible strategies for dividing spaces. In summary, by adjusting the relationship between $w$ and $w'$, the IC structure can not only retain the representation ability of the traditional neuron, but also flexibly segment linear representation spaces on complex tasks.


	


	\subsection{Application on CNN}
	The previous section discussed why the network with the IC structure works. In this section, we apply the IC structure to the convolutional structure. To simplify the notation, the activation operator is omitted. The output feature $\mathbf{u}_{i} \in \mathbb{R}^{H \times W}$ of standard convolution is given by: 
	\begin{equation}
		\begin{aligned}
			\mathbf{u}_{i} = \mathbf{w}_{i}*\mathbf{X}, \\
		\end{aligned}
	\end{equation}
	where $\mathbf{X} \in \mathbb{R}^{H' \times W' \times C'}$ is the input feature, $\mathbf{w}_{i} \in \mathbb{R}^{k \times k \times C'}$ is the set of filter kernels and $*$ is used to denote the convolutional operator. To apply the IC structure to the convolutional layer, we replace the convolutional kernel $\mathbf{w}_{i}$ with an IC kernel $[\mathbf{w}_{i}, w'_{i}]$, 
	\begin{equation}
		\label{eq9}
		\begin{aligned}
			\mathbf{u}_{i} &= [\mathbf{w}_{i}, w'_{i}]*X\\
			&= \mathbf{w}_{i}*X + \sigma(\mathbf{w}_{i}*X - w'_{i} \times (\mathbf{I}*\mathbf{X})), \\ 
		\end{aligned}
	\end{equation}
	where $\mathbf{I}$ is an all-one tensor with the same size as $\mathbf{w}_{i}$. The input feature can contain hundreds of channels and Eq. 9 will mix different features in each channel with the same proportion. Therefore, we distinguish different features by the grouped convolutional trick, 
	\begin{equation}
		\label{eq10}
		\begin{aligned}
			\mathbf{u}_{i} = \mathbf{w}_{i}*\mathbf{X} + \sigma(\mathbf{w}_{i}*\mathbf{X} - (\mathbf{I}**\mathbf{X})\mathbf{w'}_{i}), 
		\end{aligned}
	\end{equation}
	where $**$ denotes the depthwise separable convolution \cite{chollet2017xception} which separates one kernel into $C'$ parts, and $\mathbf{w'}_{i}$ is a vector with size $C'$. We term the structure described in Eq. 10 the ``IC layer". The item $\mathbf{I}**\mathbf{X}$ is termed ``rough feature", since it provides an approximate distribution of the pixels in the feature map, which helps the filters to learn more fine-grained features before the training process converges. Moreover, using a depthwise separable convolution selectively discriminates the informative features (channels) between less useful ones. Similar to the kernel method \cite{zoumpourlis2017non}, the IC layer adds non-linearity to convolution. Besides, the IC layer brings only a small increase on the computational complexity. We will show the detail in Section 3.4.

	We have mentioned that the IC structure will not change the size of the input and output, making it very applicable to a majority of CNNs. Here we list different combining strategies and show how the IC networks work. First, the simplest strategy is to replace the convolutional layers with the IC layers without changing others. In addition, it is unnecessary to implement the IC structure on a $1 \times 1$ convolution, since the rough feature of the $1 \times 1$ receptive field is the pixel itself, making it the same as a $1 \times 1$ depthwise separable convolution. 

	The second strategy is to build a block structure using the rough feature. We term this block structure the ``IC block". ResNet proposes two kinds of blocks which are popular for later network design. The basic block with two $3 \times 3$ convolutional layers is used when the number of the total layers is less than $50$. The bottleneck block which contains $1 \times 1, 3\times 3, 1\times 1$ convolutional layers are used when the number of layers is more than $50$. The structures of the IC blocks are shown in Fig. 2. Different from the first strategy, the rough feature in the IC block is transmitted across the whole block. In detail, the rough feature $\mathbf{I}**\mathbf{X}$ is calculated from the input of the block and is combined with the output of the last layer. Technically, batch normalization is used in the combining process in order to make the distribution of rough feature similar to the output of the last layer. In this way, we take advantage of the skip connection \cite{he2016identity} to transmit rough feature. Our experiments show that the IC block has a better performance compared to both the IC layer and the ResNets blocks. Moreover, it is more efficient and with less complexity, since each block calculates the rough feature only once. 


	\begin{figure}[t]
		\label{Imagenet}
		\centering
		\subfigure[]{
		\centering
		\includegraphics[]{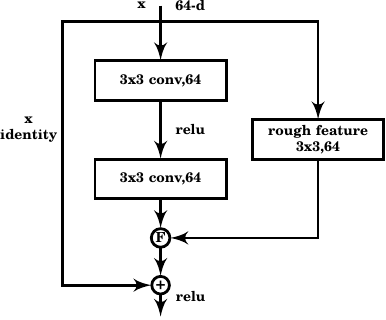}
	}
	\hspace{1.5cm}
	\subfigure[]{
		\centering
		\includegraphics[]{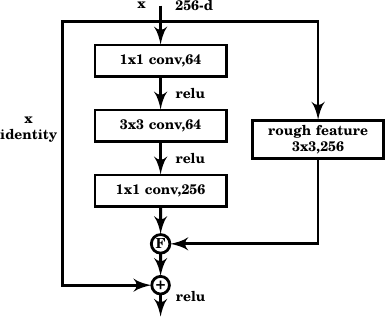}
	}
	\centering 
	\caption{IC basic block (a) and IC bottleneck block (b). $F$ denotes $F(a, b) = a + \sigma(a + b)$, where $a$ is the output of last layer and $b$ is the rough feature. The BN operator is implemented on the term $(a + b)$. } 
	\label{fig2}
	\end{figure}

	\subsection{Parameters and Complexity Analysis}\label{sec3_4}
	For the standard convolutional layer with $k \times k$ receptive field, the transformation
	\begin{equation}
		\begin{aligned}
			\mathbf{X} \in \mathbb{R}^{H' \times W' \times C'} \xrightarrow{conv} \mathbf{U} \in \mathbb{R}^{H \times W \times C}\\
		\end{aligned}
	\end{equation}
	needs $k \times k \times C' \times C$ parameters. In the IC layer, we calculate the sum of each channel without additional parameters. The weight $\mathbf{W'} = [\mathbf{w'}_{1},\mathbf{w'}_{2},\cdots,\mathbf{w'}_{C}]$ adds $1 \times 1 \times C' \times C$ parameters. Therefore, the number of parameters added by the IC layer is only $\frac{1}{k \times k}$ of the original layer. 

	The IC layer adds a depthwise separable convolution and a learnable weight $\mathbf{W'}$ which can be regarded as $1 \times 1$ convolution. The depthwise separable convolution is used to calculate the element-wise sum of the input by an all-one kernel. The increased computational complexity is same as adding a convolutional filter because we only need to do this operation once. Therefore, the increased complexity is $\frac{1}{C}$ of the original layer. The weight $\mathbf{W'}$ is a $1 \times 1$ convolution which uses less than $\frac{1}{k \times k}$ computation of the $k \times k$ convolution \cite{sifre2014rigid}. We get an approximate addition in computation of $\frac{1}{C} + \frac{1}{k \times k}$.

	\section{Experiments}
	In this section, different IC architectures will be evaluated on the task of CV, including image classification, object detection and semantic segmentation. We investigate the effectiveness of the IC structure by a series of comparative experiments detailed below.

	\begin{figure}[t]
		\label{Imagenet}
		\subfigure[]{
		\centering
		\includegraphics[scale = 0.35]{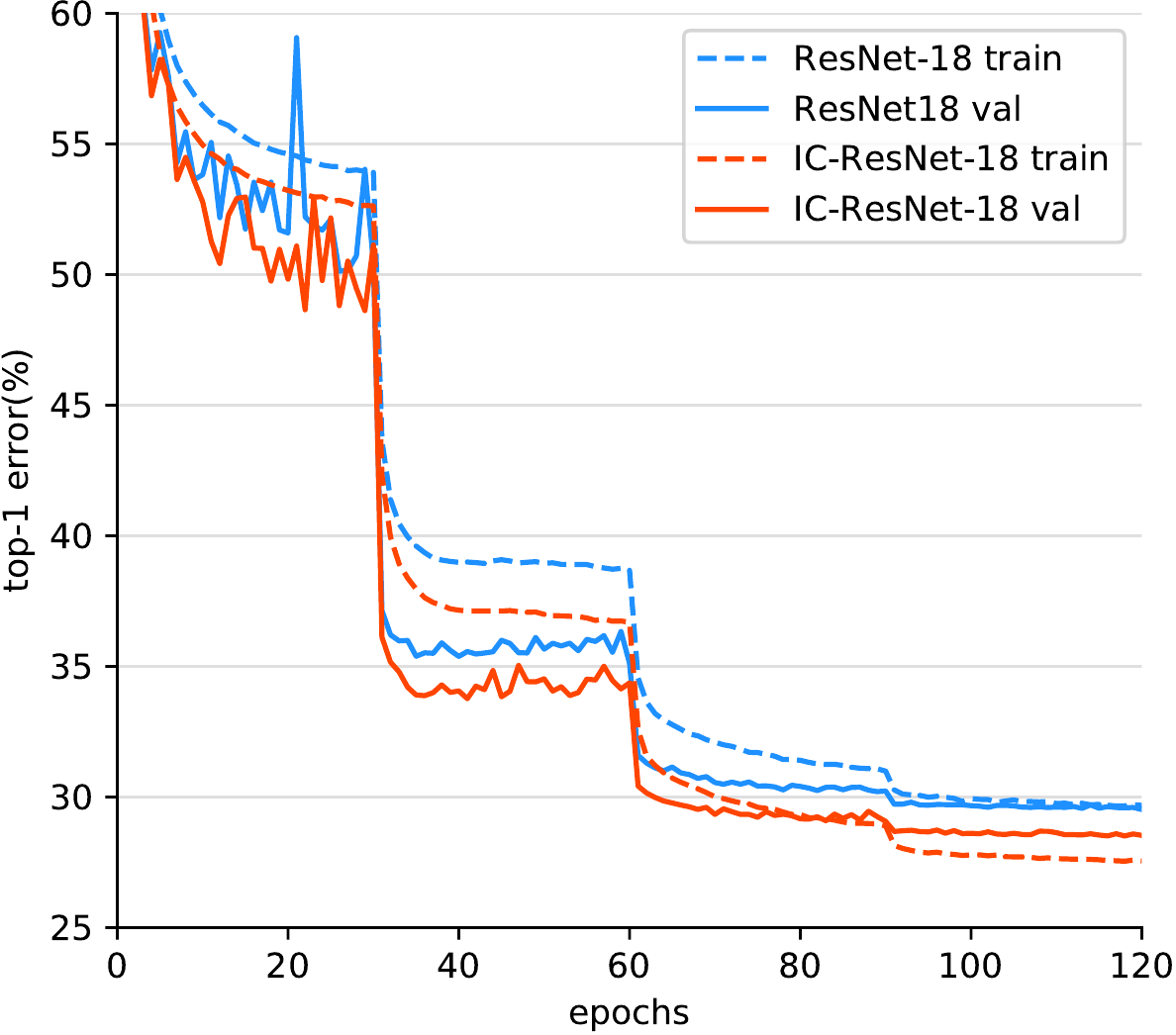}
		}
		\subfigure[]{
		\centering
		\includegraphics[scale = 0.35]{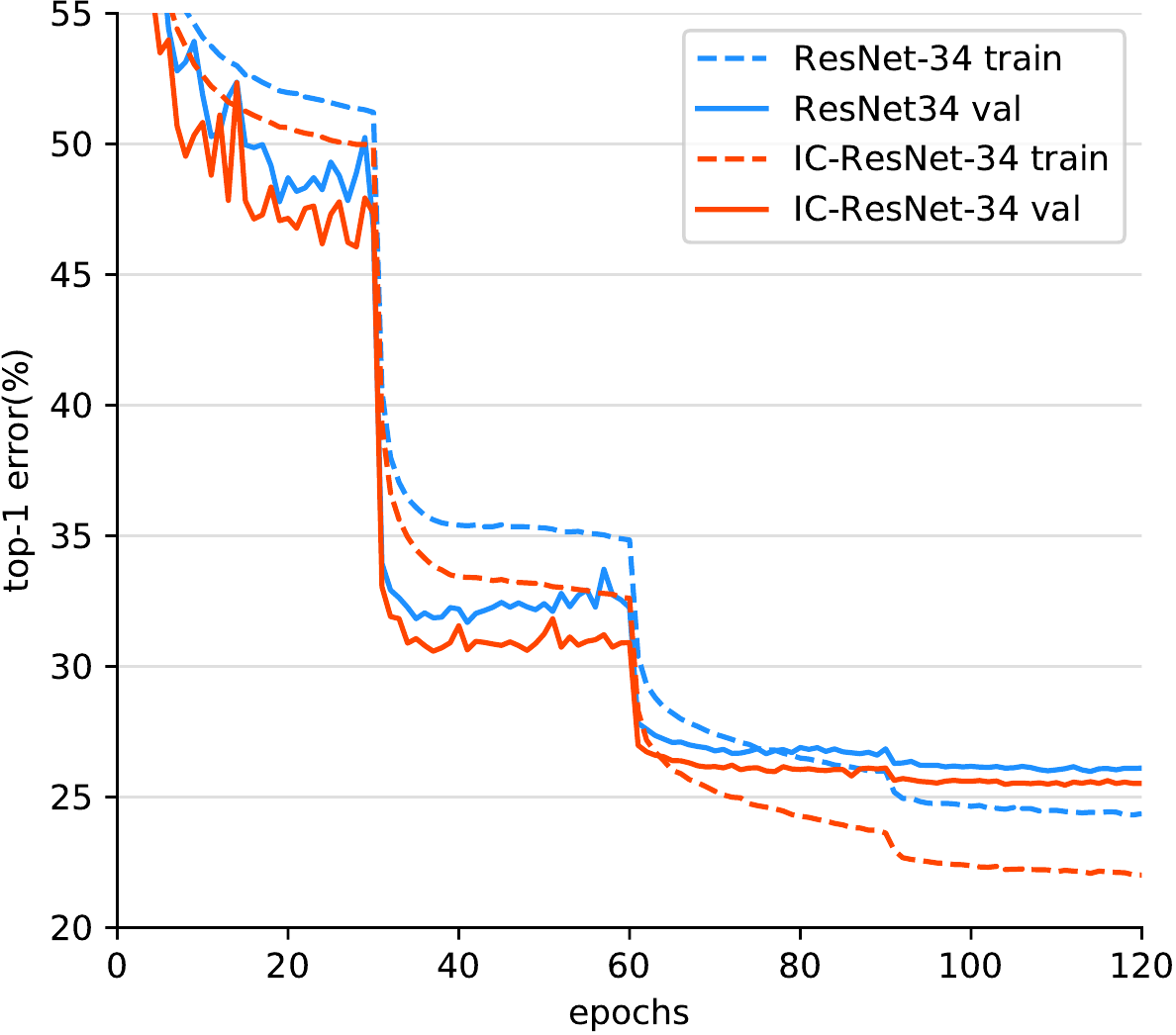}
		}
		\subfigure[]{
		\centering
		\includegraphics[scale = 0.35]{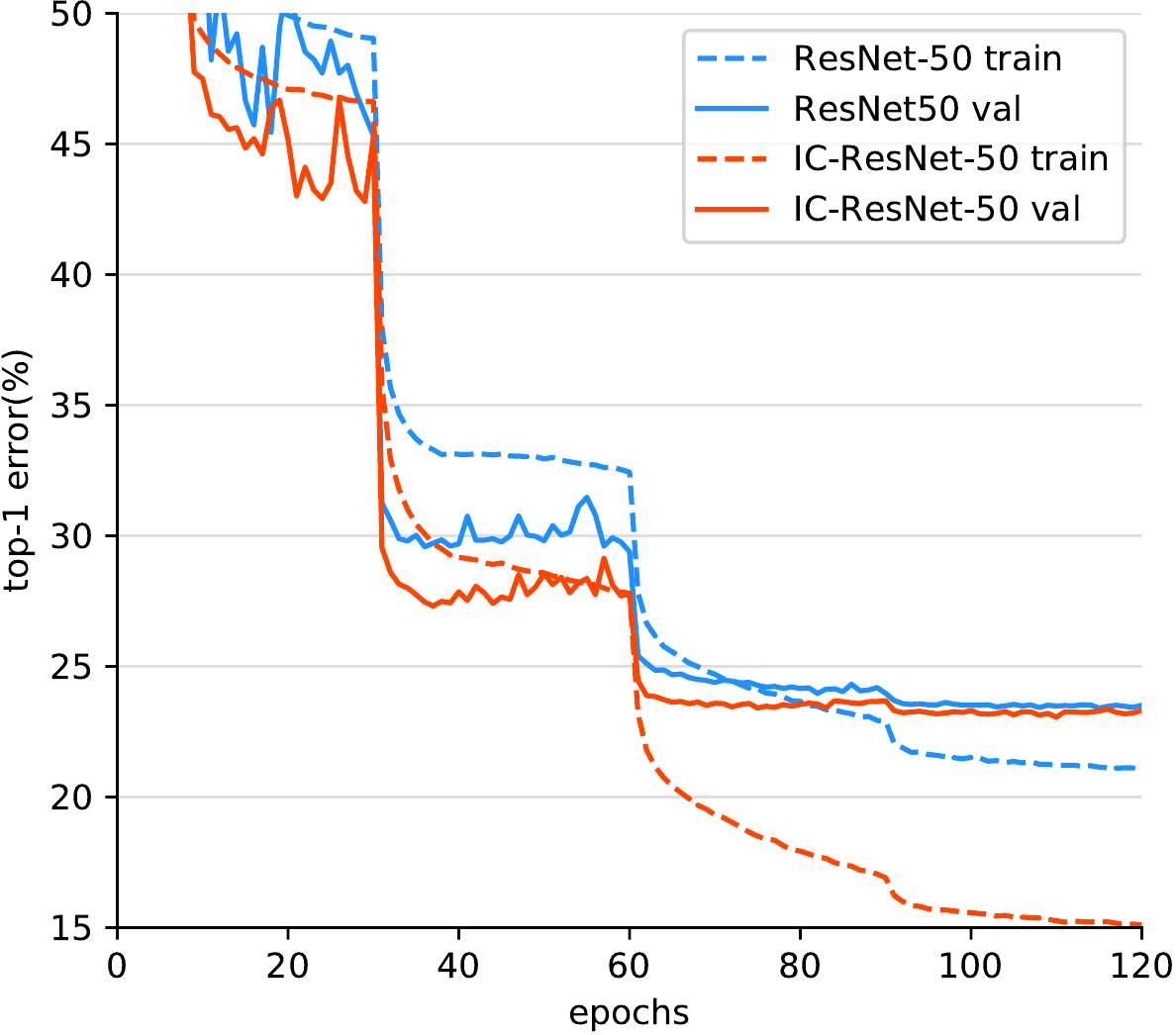}
		}
		\caption{Training curves of the three IC-ResNets and their basic models on ImageNet. Dashed lines denote training error, and solid lines denote validation error of the center crops. IC-ResNets exhibit improved performance both on convergence speed and error rate.} 
		\label{fig3}
	\end{figure}

	\begin{table}
		\caption{The error rate of models on ImageNet. The B versions represent using the IC layer on basic model. ResNets with the superscript * represent the results reported in previous work.}
		\begin{tabular}{l|cc|c}  
		\toprule
		Model  & Top-1(single-crop /10-crop) & Top-5(single-crop/10-crop) & GFlops \\
		\midrule
		ResNet-18       & 29.52 / 27.30 & 10.29 / 8.92 & 1.82  \\
		IC-ResNet-18	& 28.51 / 26.33 & 9.72 / 8.46 & 1.91  \\
		IC-ResNet-18-B	& 28.56 / 26.69 & 9.80 / 8.56 & 2.01  \\
		ResNet-18*      & 30.24 / 28.22 & 10.92 / 9.42 & -  \\
		\midrule
		ResNet-34   	& 25.98 / 24.04 & 8.21 / 7.12 & 3.68  \\
		IC-ResNet-34	& 25.45 / 23.62 & 8.04 / 6.92 & 3.86  \\
		IC-ResNet-34-B	& 25.55 / 23.49 & 7.90 / 6.86 & 4.07  \\
		ResNet-34*  	& 26.70 / 24.52 & 8.58 / 7.46 & -  \\
		\midrule 
		ResNet-50   	& 23.55 / 22.38 & 6.95 / 6.54 & 4.12  \\
		IC-ResNet-50	& 23.05 / 21.49 & 6.74 / 5.89 & 6.96  \\
		IC-ResNet-50-B	& 23.40 / 21.96 & 6.81 / 5.99 & 4.33  \\
		ResNet-50*		& 23.85 / 22.85 & 7.13 / 6.71 & -  \\
		\bottomrule
		\end{tabular}
		\label{tab1}
	\end{table}

	\textbf{ImageNet.} The ILSVRC 2012 classification dataset \cite{russakovsky2015imagenet} consists of more than 1 million colour images in 1000 classes divided into 1.28M training images and 50K validation images. We use three versions of ResNet (ResNet-18, ResNet-34, ResNet-50) to build the corresponding IC networks. For fair comparison, all our experiments on ImageNet are conducted in the same environment setting. The optimizer uses stochastic gradient descent (SGD) \cite{lecun1989backpropagation} method with a weight decay of $10^{-4}$ and a momentum of $0.9$. The training process is set to $120$ epochs with a batch size of $256$. The learning rate is set to $0.1$ at the beginning and will be reduced 10 times every $30$ epochs. Besides, all experiments are implemented with the Pytorch \cite{paszke2019pytorch} framework on a server with 4 NVIDIA TITAN Xp GPUs. To remove the impacts of different environments, we also trained the original ResNet models reported in Table 1 with the same configuration.
	
	We apply both the single-crop and 10-crop for testing. Both IC-ResNet-18 and IC-ResNet-34 are constructed using the same basic IC block structure (Fig. 2(a)). Their training curves are depicted in Fig. 3(a, b). We observe that both IC networks have faster convergence speed at each learning stage. The validation error and FLOPs are reported in Table 1. We observe that the IC-ResNet-18 and IC-ResNet-34 can obviously reduce the 10-crop top-1 error by $0.97\%, 0.42\%$ and the top-5 error by $0.46\%, 0.20\%$ with a small increase in the calculation ($4.95\%, 4.89\%$), validating the effectiveness of the IC basic block. The IC-ResNet-50 is constructed using the bottleneck IC block (Fig. 2(b)). Its training curve is shown in Fig. 3(c). The 10-crop top-1 error is $21.49\%$ and the top-5 error is $5.89\%$, exceeding ResNet-50 by $0.89\%$ and $0.65\%$ respectively. This result also exceeds the error rate achieved by the deeper ResNet-101 network ($21.75\%$ top-1 and $6.05\%$ top-5 error) with fewer FLOPs ($6.96$ GFLOPs vs. $7.85$ GFLOPs). For the deeper ResNets, we argue that the deeper IC-ResNets can get similar results, since they all use the same bottleneck block as ResNet-50. 

	In Table 1, the results from ResNets are from \cite{he2016deep} \cite{he2016identity} and \cite{wang2019kervolutional}, which are slightly lower than results reproduced by us. This is mainly because we add an extra stage using the learning rate $0.0001$. Besides, we evaluate the networks using the IC layers to replace the convolutional layers. Although these networks called "IC-ResNet-B" can not achieve the same performance of IC-ResNets in most circumstances, they can reduce the 10-crop top-1 error by $0.61\%, 0.55\%, 0.42\%$ compared to the corresponding ResNets. The final subsection will analyze the difference between the two kinds of IC networks in detail.

	\begin{figure*}[t]
		\label{Imagenet}
		\subfigure[]{
		\centering
		\includegraphics[scale = 0.31]{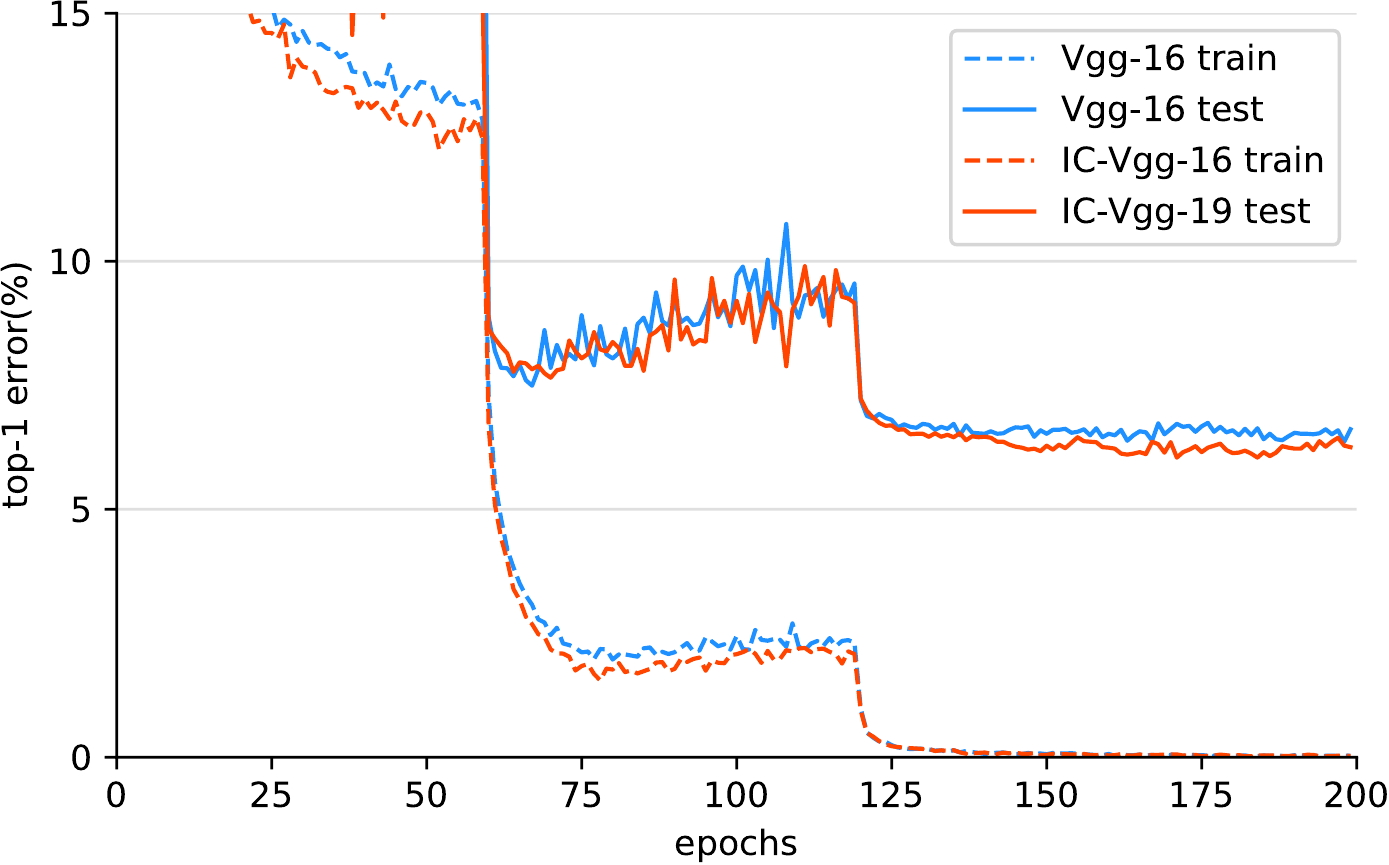}
		}
		\subfigure[]{
		\centering
		\includegraphics[scale = 0.31]{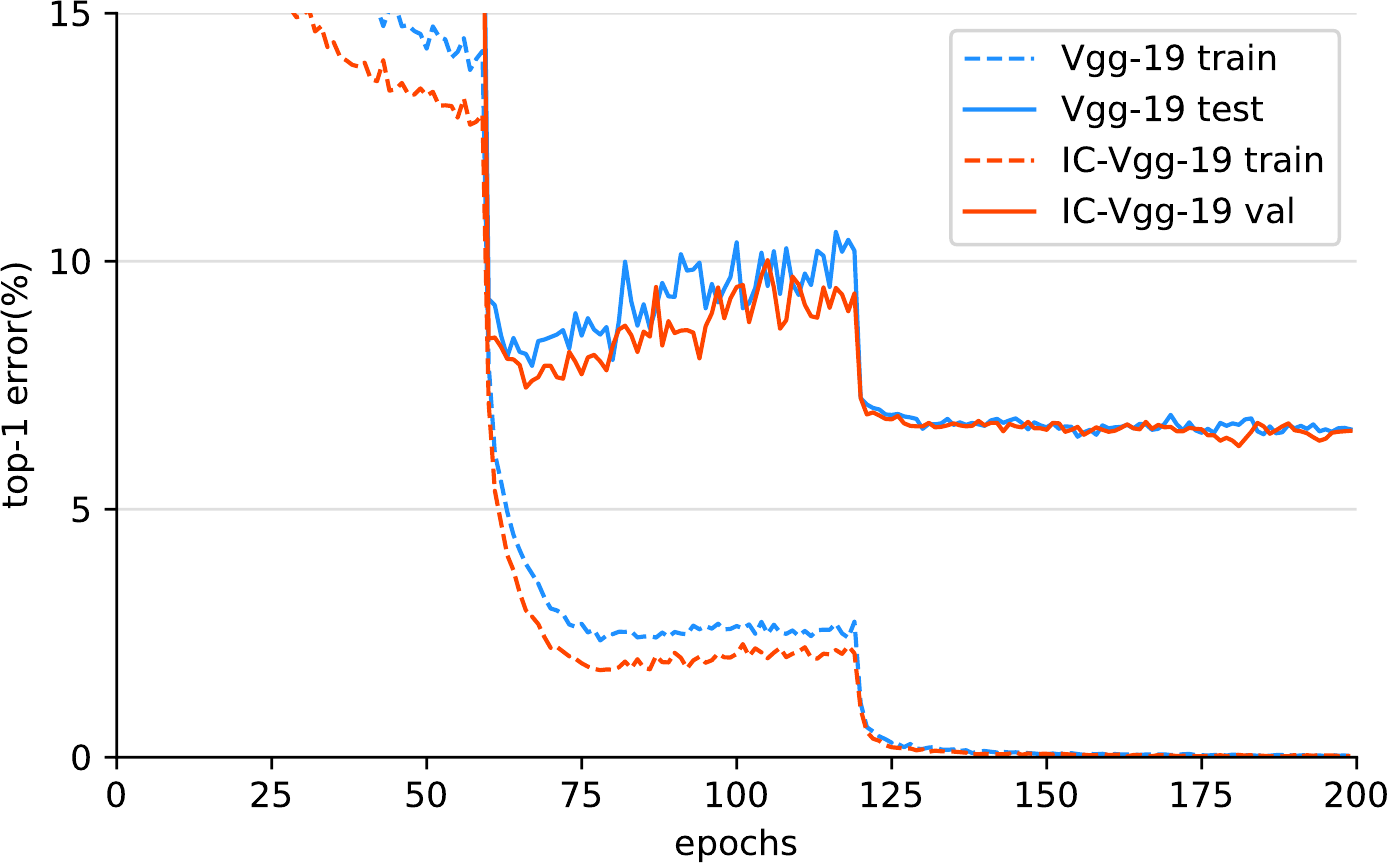}
		}
		\subfigure[]{
		\centering
		\includegraphics[scale = 0.31]{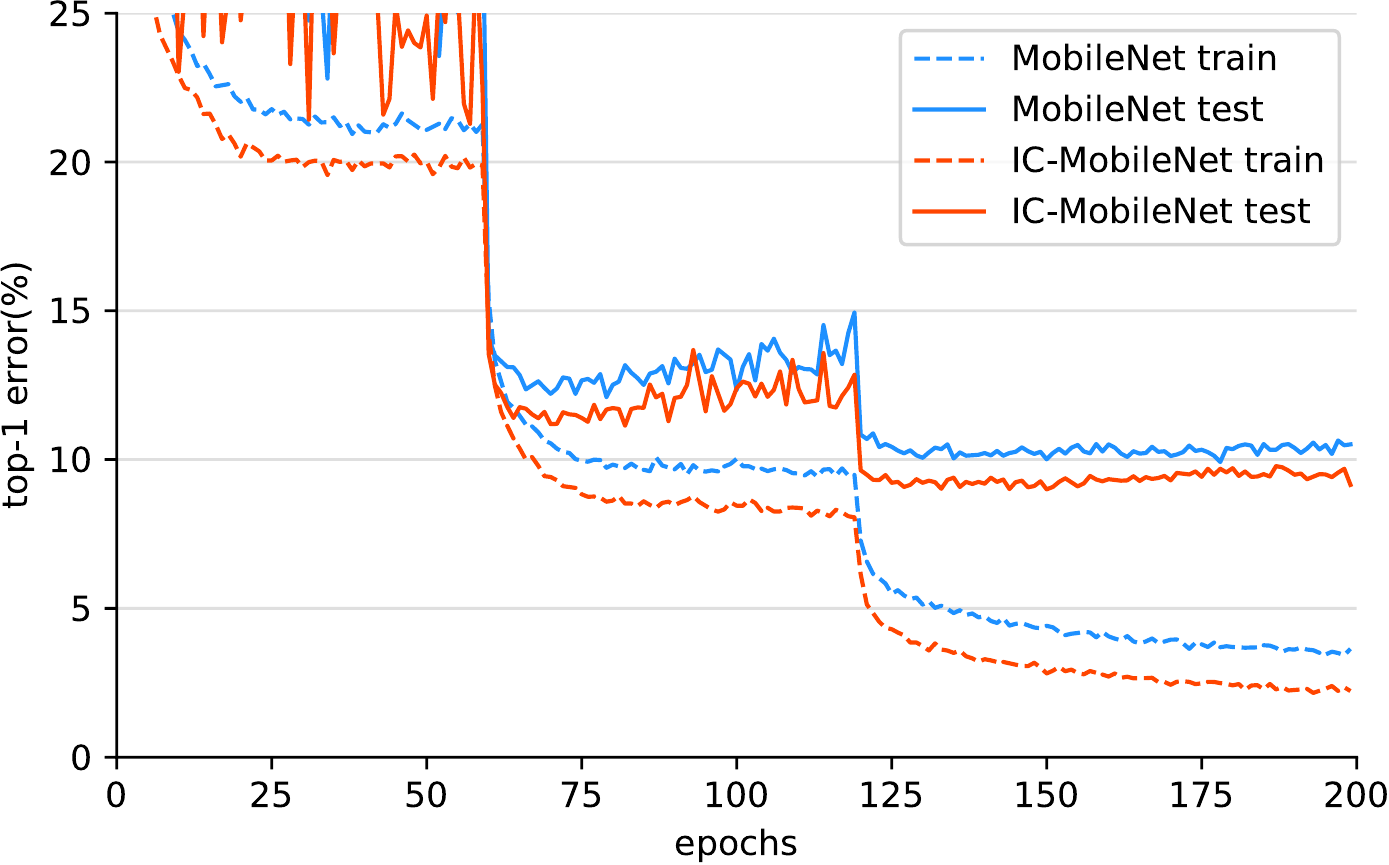}
		}
		\subfigure[]{
		\centering
		\includegraphics[scale = 0.31]{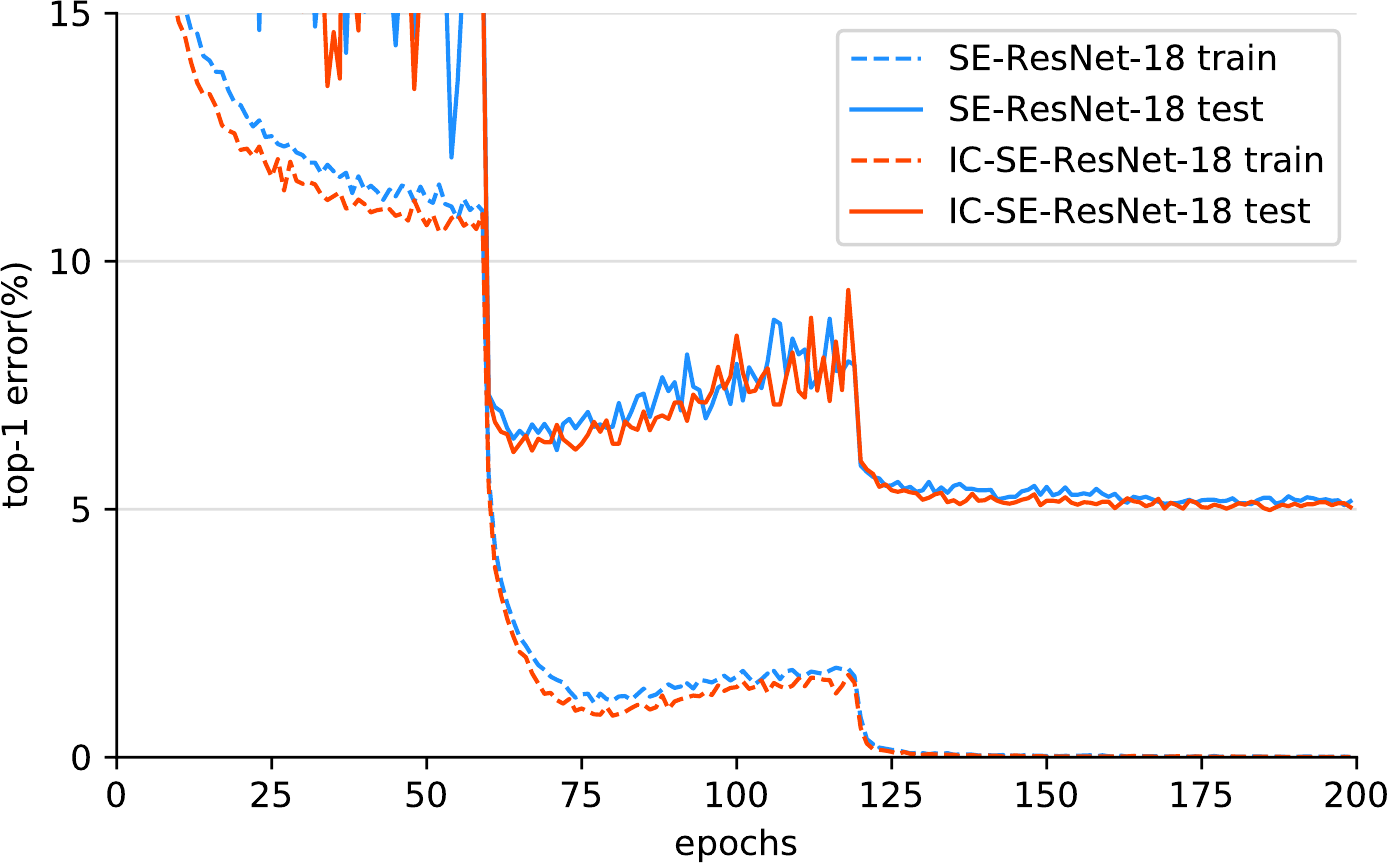}
		}
		\subfigure[]{
		\centering
		\includegraphics[scale = 0.31]{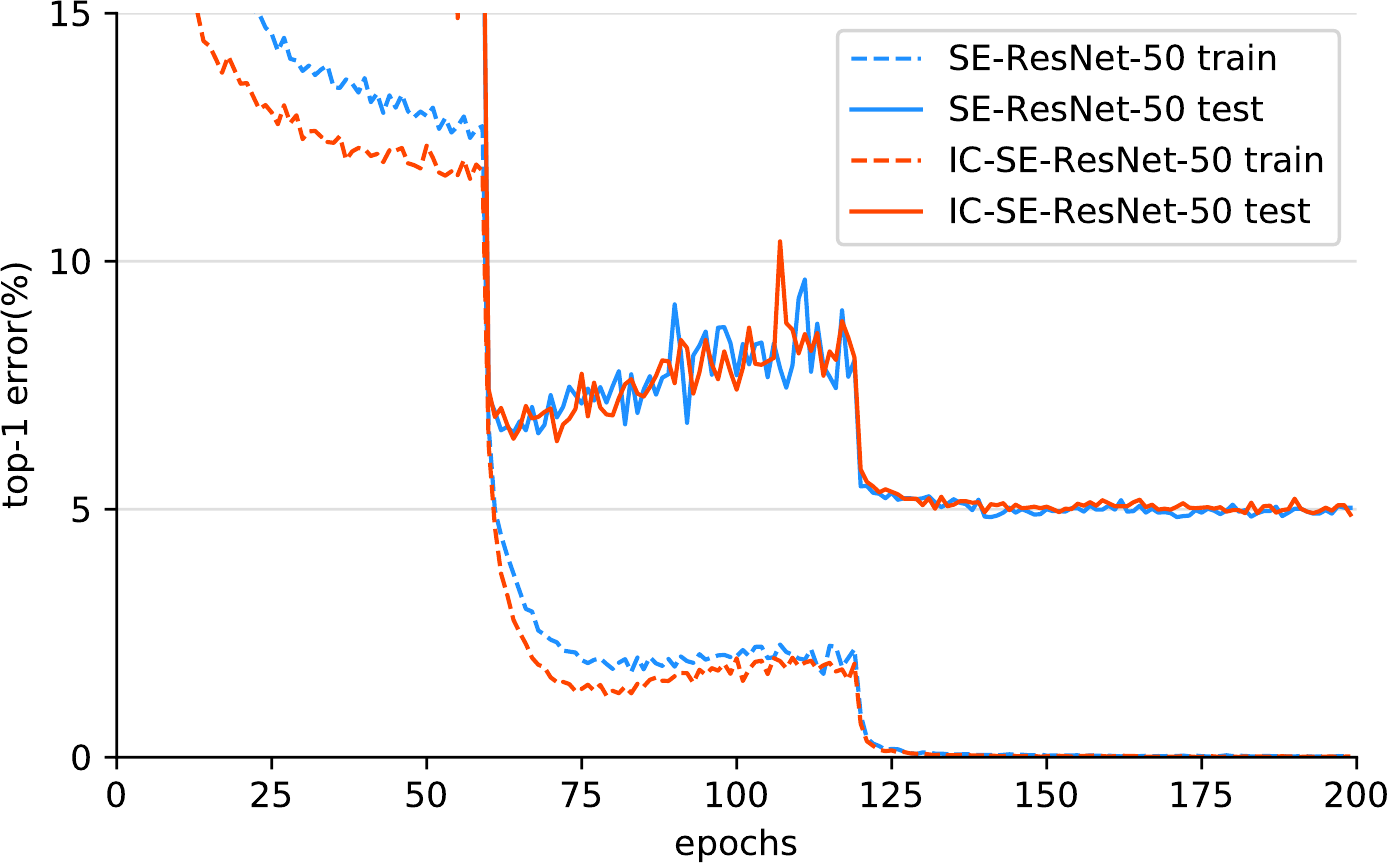}
		}
		\subfigure[]{
		\centering
		\includegraphics[scale = 0.31]{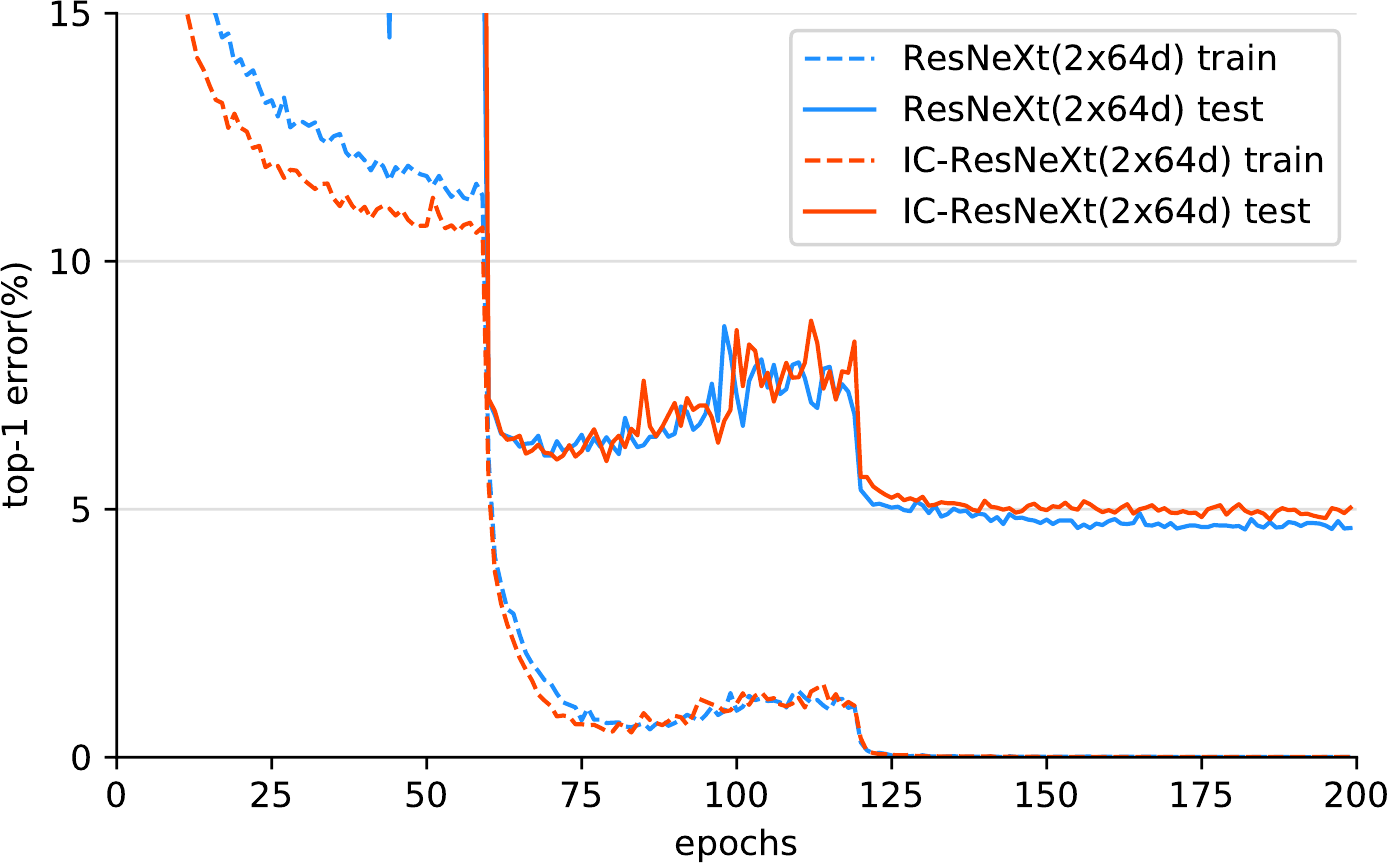}
		}
		\caption{Training curves of the six IC networks and their basic models on CIFAR10. The IC networks use the IC layer in (a), (b), (c), and use the IC block in (d), (e), (f).} 
		\label{fig4}
	\end{figure*}

	\begin{table}
		\caption{The top-1 error of models on CIFAR10. We use our environment settings to reprodurce the baseline results.}
		\centering
		\begin{tabular}{lc|lc|lc}  
		\toprule
		Model  & Top-1 & Model & Top-1 & Model & Top-1\\
		\midrule
		VGG-16      & 6.36  & VGG-19 & 6.46 & MobileNet & 9.92 \\
		IC-VGG-16	& 6.04  & IC-VGG-19 & 6.27 & IC-MobileNet & 9.00  \\
		\midrule
		SE-ResNet-18 & 5.08     & SE-ResNet-50 & 4.84 & ResNext(2x64d) & 4.59 \\
		IC-SE-ResNet-18 & 4.85  & IC-SE-ResNet-50 & 5.05 & IC-ResNext(2x64d) & 4.62\\
		\bottomrule
		\end{tabular}
		\label{tab2}
	\end{table}

	\textbf{Combining with modern models.} We further investigate the universality of the IC structure by integrating it into some other modern architectures. These experiments are conducted on the CIFAR10 dataset, which consists of 60K $32 \times 32$ colour images in 10 classes divided into 50K training images and 10K testing images. Each model is trained in $200$ epochs with a batch size of $128$. The learning rate is initialized to $0.1$, which will be reduced $10$ times at the $60$th epoch and the $120$th epoch. The optimizer settings are the same as in the ImageNet experiment. 


	We integrate the IC structure into VGGNet(VGG-16 and VGG-19 versions) and MobileNet \cite{howard2017mobilenets} replacing the convolutional layers with the IC layers. Specially, the adjacent convolution layers (a depthwise separable convolution layer and a pointwise convolutional layer) in MobileNet are treated as one convolution layer when integrating with the IC layer because there is a close relationship between adjacent layers. Fig. 4(a, b, c) shows the training curves and top-1 error is listed in Table 2, we observe that the IC layers can improve the convergence speed and generalization abilities networks.

	We apply the IC structure to SENet (SE-ResNet-18 and SE-ResNet-50 versions) and ResNeXt(we choose the 2x64d version) by replacing their block structures with the IC blocks. It is worth noting that SENet block structure has two components, one is the ResNet block, the other is the squeeze-and-excitation (SE) operator. We only replace the ResNet block component with the IC block in each SENet block. As for ResNeXt, it uses the bottleneck block as the building block. The property that each $3 \times 3$ convolutional layer in the bottleneck blocks is separated is kept when replacing the bottleneck blocks with the IC blocks. The training curves of IC-SE-ResNet-18, IC-SE-ResNet-50, and IC-ResNeXt are depicted in Fig. 4(d, e, f) and the top-1 error is listed in Table 2. We notice that IC-SE-ResNet-18 has lower test error than SE-ResNet-18, while both the IC-SE-ResNet-50 and IC-ResNeXt show no improvement.

	\textbf{Overfitting.} In ImageNet experiments, we observed that the training accuracy of IC-ResNet-50 is much higher than the validation accuracy. We think this is caused by overfitting. In the CIFAR10 experiment, the results of IC-SE-ResNet-50 and IC-ReNeXt also reflect this problem: their test accuracy is lower than the corresponding basic network while the train accuracy of both is approximately $100\%$. When there is sufficient training data, overfitting generally indicates that the model complexity is too high or too low. In our experiments, we believe that the overfitting is caused by the high complexity of the model. The IC structure will not hurt the non-linear fitting ability of the basic network. and overfitting appears because the IC networks generate overly complex classification boundaries. In Fig. 4(e, f), although the IC-SE-ResNet-50 and IC-ResNeXt have no improvements on final accuracy, they obviously exceed the basic models when not fitting well (when learning rates are $0.1$ and $0.01$). 

	In general, there are some implicit distribution gaps between the training set and the test set, which cause the trained model to fail to achieve the accuracy of the training set on the test set. In our experiments, the obvious overfitting phenomenon only exists in IC-resnet50, IC-renext, and IC-senet50. These three models use similar bottleneck blocks. We assume that the IC bottleneck block enlarged the the gap between training set with validation set (test set). To further confirm our hypothesis, we trained IC-ResNet-101 and ResNet-101 with different settings and equipment compared to ImageNet experiments. Although the single crop top-1 error of IC-ResNet-101 ($22.17\%$) exceeds the original ResNet101 ($22.63\%$), the gap between two datasets is increased from $2.35\%$ to $6.79\%$. This experiment comfirms that the IC bottleneck block increases the risk of overfitting.



	\textbf{The applications of the IC layer and IC block.} Table 1 shows the comparison between networks with the IC blocks and their B versions with the IC layers. Both two structures have an improvement on their basic networks while the models with the IC block achieve the best results (except the 10-crop results on 34-layer architectures). We argue that this is because the rough feature benefits from the skip connection which solves the degradation problem and makes the gradient of the rough feature obvious. Higher performance makes the IC blocks a greater choice when they are applied in the model with a similar block structure. Besides, the IC layer has the characteristic of a smaller structure, making it more flexible to integrate into the majority of models.


	\section{Conclusion and Future Work}

	Inspired by the elastic collision model, we propose the IC structure that could be used to improve the performance of convolution networks. We build the IC networks by integrating the IC structure into the state-of-the-art models. Our experiments show that the IC networks could effectively improve the convergence speed with a little additional computational burden. Besides, our IC network reaches a higher accuracy compared to the baseline on the image classification task.

	In this work, we have not fully explored the applicability to other networks that the IC structure potentially enables. Our future work includes applications to other architectures and tasks beyond image classification.

\section*{Broader Impact}

We have proposed a new calculation unit called the ``IC'' structure, and analyzed its advantages and disadvantages in detail. The results in the our paper can be easily reproduced. We consider the broader impact of our work: First, our work will benefit researchers who use CNN to extract features. Second, we propose a new method that may provide more flexible ideas for designing nonlinear units. However, the IC structure may increase the overfitting risk, which will affect the performance of models interleaved by some complex networks in the validation set or test set. Besides, we do not leverage biases in the data.

%

\small

\bibliographystyle{plainnat}
\bibliography{oldrefer}

\end{document}